%% file: ms.tex
\documentclass{article}
\usepackage{hyperref}
\usepackage[ruled]{algorithm2e}
\usepackage{amsmath}
\usepackage{amsfonts}
\usepackage{amssymb}
\usepackage{multirow}
\usepackage{float}

\usepackage{subfig}
\usepackage{graphicx}
\usepackage{xcolor}  

\usepackage{siunitx}
\usepackage{booktabs}
\usepackage{csvsimple}
\usepackage{authblk}
\usepackage{hyperref}
\usepackage{float}

\SetKw{Continue}{continue}

\SetKw{Or}{ or }
\SetKw{Break}{break}

\DeclareMathOperator*{\argmax}{arg\,max}

\newcommand{\own}{\textsc{optisplit}\xspace}
\newcommand{\sss}{\textsc{SS}\xspace}
\newcommand{\is}{\textsc{IS}\xspace}
\newcommand{\sois}{\textsc{SOIS}\xspace}
\newcommand{\pmbsrs}{\textsc{PMBSRS}\xspace}
\newcommand{\random}{\textsc{Random}\xspace}

\newcommand{\delicious}{\textsc{delicious}\xspace}
\newcommand{\mediamill}{\textsc{mediamill}\xspace}
\newcommand{\bibtex}{\textsc{bibtex}\xspace}
\newcommand{\cc}{\textsc{CC}\xspace}
\newcommand{\mf}{\textsc{MF}\xspace}
\newcommand{\bp}{\textsc{BP}\xspace}
\newcommand{\go}{\textsc{GO}\xspace}
\newcommand{\wiki}{\textsc{Wiki10-31K}\xspace}

\newcommand{\ld}{\textsc{LD}\xspace}
\newcommand{\wld}{\text{rLD}\xspace}
\newcommand{\ed}{\textsc{ED}\xspace}
\newcommand{\dcp}{\textsc{DCP}\xspace}

\newcommand{\bigohpar}[1]{\ensuremath{\mathcal{O} \left(#1\right)}}

\title{Novel split quality measures
for stratified multilabel Cross Validation with
application to large and sparse gene ontology
datasets}

\author[1]{Henri Tiittanen}
\author[1,2]{Liisa Holm}
\author[1]{Petri Törönen}

\affil[1]{Institute of Biotechnology, Helsinki Institute of Life Sciences (HiLife), University of Helsinki, Helsinki, Finland}
\affil[2]{Organismal and Evotionary Biology Research Program, Faculty of Biosciences, University of Helsinki}

\begin{document}

\maketitle

\section{Abstract}
Multilabel learning is an important topic in machine learning research.
Evaluating models in multilabel settings requires specific cross validation methods designed for multilabel data. In this article, we show that the most widely used cross validation split quality measures do not behave adequately with multilabel data that has strong class imbalance. We present improved measures and an algorithm, \own, for optimising cross validations splits. We present an extensive comparison of various types of cross validation methods in which we show that \own produces more even cross validation splits than the existing methods and that it is fast enough to be used on big Gene Ontology (\go) datasets.

\section{Introduction}\label{introduction}

Cross validation is a central procedure in machine learning, statistics and other fields. It evaluates model performance by testing models on data points, excluded from training dataset. In standard cross validation a dataset $D$ is split randomly into $k$ non overlapping subsets (folds) $D_i, i\in k$. A model is trained for each fold $i$ on the data $\cup_{j\in k| j \neq i} D_j$ and evaluated on $D_i$. So one subset, $D_i$, is left out of the training process and used as an evaluation data. The averaged result over all the folds represents the final performance. Cross validation is typically used when the amount of data is limited. When abundant amount / very large amounts of data is available, one can also use standard training test splits in place of cross validation.
Cross validation is further used, besides traditional model evaluation, also when various predictive models are combined in classifier stacking \cite{wolpert} and when various parameters in predictive models are optimized.
The typically used random split approach assumes that the positive and negative class distributions are balanced. If the class distributions are imbalanced the resulting splits may not allow efficient learning. As an example, suppose that in binary classification settings one randomly generated fold contains all the data points in the data belonging to one of the classes. Then the corresponding training set consisting of the rest of the folds does not contain any data points of that class and the model cannot learn anything about the class.

\begin{figure}
    \centering
    \includegraphics[width = 1.0\textwidth]{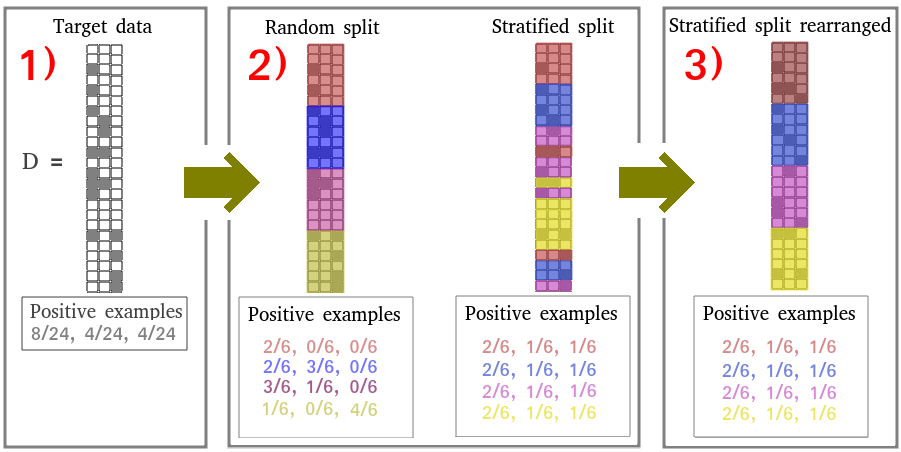}
    \caption{Examples of stratified multilabel cross validation with splits of different quality. 1) presents a target data with 24 data points and three classes. This could be a subset of a sparse high dimensional dataset.
    2) Now, the data is splitted into four cross validation folds in two ways, one representing a random split and the other representing a stratified split optimized over all classes. The ideal distribution of the positive data points in each fold would be 2/6 for the first class and 1/6 for the second and third class so that they would follow the class distributions of the whole data. Random split results in very unbalanced folds while the stratified split follows closely the data distribution for all classes.
    3) Here, the well balanced stratified split is rearranged for increased clarity.
    }
    \label{fig:my_label}
\end{figure}

Stratified cross validation methods are variants of cross validation that ensure that the class distributions of the folds are close to the class distributions of the whole data. With class imbalanced data these stratified methods ensure especially the distribution of smaller classes. In this work we focus on stratified cross validation applied to multilabel classification. Multilabel classification presents additional challenges for stratified cross validation as each data point can belong to multiple classes simultaneously (see figure \ref{fig:my_label}). Here each class represents a separate task for training and evaluation. The cross validation split should be formed so that the correct class distributions are maintained for all classes and all folds at the same time. This ensures that the classifier to be evaluated can be trained and tested effectively with all of the created data splits. Random splitting has been historically a popular choice on multilabel data but is has been shown to lead to poor results \cite{sechidis11}.

As dataset sizes are growing across application domains, multilabel and extreme classification are of growing interest and importance \cite{Bengio19}. Thus, assessing model quality on these settings is getting more and more important. Furthermore, the model assessment is critical step when different predictions are combined in various classifier stacking methods \cite{wolpert}.
However, we show in Section~\ref{ssec:ems} and Section~\ref{sec:cvexp} that the current stratified cross validation methods use suboptimal split quality measures.  Also, many existing methods are impractically slow for extremely big classification datasets. Here we present a new algorithm for generating multilabel cross validation datasets based on optimising the global distribution of all classes.

The methods presented here are developed in the context of gene ontology (\go) data \cite{ashburner2000go} for automated protein function prediction task \cite{zhou2019cafa}. The \go datasets used here are high dimensional and contain over half a million data points. The high number of classes, often in the order of thousands to tens of thousands, presents a challenge for analysis. Furthermore, the classes have a hierarchical tree structure, which causes the class distributions to be strongly imbalanced. So, as a result, most classes contain a very small number of data points while some classes have a very large number of data points. The high number of small classes, i.e. classes that have few positive data points, makes the negative data abundant for most of the classes and results in difficulties to have enough positive data points to train the models.

This article is organised as follows. In Section~\ref{related}, we review the related work. In Section~\ref{metrics}, we show that the currently most widely used stratified multilabel cross validation split quality measure is inadequate. We present new better measures and demonstrate the behaviour of these first with synthetic data splits. In Section~\ref{sec:algorithm}, we present an algorithm for optimising cross validation splits with respect to any selected split quality measures. In Section~\ref{sec:cvexp}, we present a comparison of the algorithms on a wide range of real-world datasets and conclude that the new algorithm is the best practical choice for getting balanced splits as measured by the new measures. Finally, we conclude with a discussion and present ideas for future work in Section~\ref{discussion}.

\section{Related work}\label{related}

The most widely known method for generating cross validation splits with
balanced class distributions is Iterative Stratification (\is)
\cite{sechidis11}. \is works by dividing the
data points evenly into the folds one class at a time. It always chooses the class with the fewest positive data points for the processing, and breaks ties first by the largest number of desired data points and further randomly. Smaller classes are more difficult to balance equally among the folds so starting from them makes sure they get well distributed. Bigger classes are easier to distribute so distributing them later is justifiable. Iterative stratification has also been extended to consider second order relationships between labels. This method is known as Second Order Iterative Stratification (\sois) \cite{szymanski17}.

The recently introduced Stratified Sampling (\sss) algorithm \cite{merrillees21} is designed to produce balanced train/test splits for extreme classification data with a high number of data points and dimensions. It should be faster to use than iterative stratification variants and produce splits with better distributions. This method calculates the proportion of each class in training and test sets. Then for each data point a score is calculated over its positive classes and the data points with the highest scores are redistributed from one to the other partition. This method needs three parameters that have to be adjusted according to the data and it does not produce cross validation splits directly but training/test splits. 

The partitioning method based on stratified random sampling (\pmbsrs) \cite{charte16} is based on using the similarity of the label distributions between data points to group them and then dividing them into the cross validation folds \cite{charte16}. A similarity score is defined as the product of the relative frequencies of the positive labels present in each data point.  Then the data points are ordered by the score into a list which is cut into as many disjoint subsets S as is the desired number of cross validation folds.
Each cross validation fold is then generated by randomly selecting items without replacement from each set $s\in S$ so that each fold gets an equal proportion of the samples from each $s$. The end result is that each fold contains elements with different scores. As a non iterative method resembling the basic random sample method this is computationally less expensive than the iterative methods. However, this does not measure the split quality directly but is more aimed to ensure that all folds contain an equal amount of differently sized classes.
    
\section{Cross validation split quality measures}\label{metrics}

The current cross validation split quality measures are either evaluating the quality of the folds directly or applying some model on the folds and comparing the learning results. Here we focus on directly comparing the quality of the folds in order to make the comparisons model independent. The most commonly used measures in literature are the Labels Distribution (\ld) measure and the Examples Distribution (\ed) measure \cite{sechidis11}.

Let $n$ be the number of data points, $k$ be the number of cross validation folds and $q$ be the number of classes. $D \in \{0,1\}^{n\times q}$ denotes the multilabel target set and $S_j, j \in 1\dots k$ denotes the folds which are disjoint subsets of $D$. The subsets of $D$ and $S_j$ containing positive data points of label $i \in 1\dots q$ are denoted as $D^i$ and $S_j^i$

We define the positive and negative frequencies for fold $j$ and label $i$ as $p_j^i=|S_j^i|/|S_j|$ and $1-p_j^i$, respectively. Similarly for the whole data, positive frequency as $d^i=|D^i|/|D|$ and negative frequency as $1-d^i$. 

Then, we define
\begin{equation}\label{eq_ld}
\ld = \frac{1}{q}\sum_{i=1}^{q}\left(\frac{1}{k}\sum_{j=1}^{k}\left|\frac{p_j^i}{1-p_j^i} - \frac{d^i}{1-d^i}\right|\right)
\end{equation}
and  
\begin{equation}\label{eq_ed}
\ed = \frac{1}{k}\sum_{j=1}^{k}\left||S_j| - \frac{|D|}{k}\right|.
\end{equation}

Intuitively, \ld measures how the distribution of the positive and negative data points of each label in each subset compares to the distribution in the whole data \footnote{\ld compares odds. Odds := positive to negative ratio.} and \ed measures the deviation between the number of data points of each subset and the desired number of data points. Since the exact equality of the fold sizes is not generally important in practice, the \ed score is merely useful in checking that the fold size differences are sufficiently small compared to the data size. 
 
As noted in Section~\ref{ssec:ems}, it is especially important for training and evaluation of models that the distributions of small classes in the cross validation folds are well balanced. Small classes are hardest to split well, since there are fewer ways to distribute the data points and even small differences in fold distributions give big relative differences. A good split quality measure should be able to correctly quantify the quality of the folds even on small classes. 

The \ld (Equation~\ref{eq_ld}), is defined as the arithmetic mean
of the differences between the positive frequencies of a cross validation fold and of the whole data over all classes and folds, using a transformation of the form $f(x) = x/(1-x)$ for both frequencies. Since the data size is fixed, the frequencies $p^i_j$ and $d^i$ are linear functions of the class sizes $|S_j^i|$ and $|D^i|$, respectively. Since $x < x/(1-x), \forall x\in [0,1)$ and $f'(x) = 1/(1-x^2) > 1$, the transformed quantities grow faster than the class size, resulting that the difference is greater for bigger classes than it should be. That is, the same absolute difference of $p_j^i$ from $d^i$ results in a larger contribution by bigger classes.\footnote{The contribution is equal for the special case $p_j^i = d^i$ when the difference is zero}.

Hence, as an improved alternative, we propose the relative Labels Distribution (\wld) measure

\begin{equation}\label{eq_wld}
\wld = \frac{1}{q}\sum_{i=1}^{q}\left(\frac{1}{k}\sum_{j=1}^{k}\left|\frac{d^i-p_j^i}{d^i}\right|\right).
\end{equation}

Intuitively, \wld measures the linear difference of the ratios of the positive frequencies between each fold and the whole data weighted by the class size. Since the difference is calculated without the offending nonlinear transformation, the measure is class size independent, while still having the same main operating principle as \ld of comparing the distributions (frequencies) of folds to those of the whole data.

In addition, we present another measure that is insensitive to
the class size, the Delta-Class Proportion (\dcp)

\begin{equation}\label{eq_dcp}
\dcp = \frac{1}{q}\sum_{i=1}^{q}\left|\frac{\max_{j\in 1\dots k}(|S_j^i|)}{|S^i|}-\frac{1}{k}\right|,
\end{equation}

where the first part of the function represents the observed result and 1/k is the positive frequency of a flat distribution.
Compared to \wld this measure does not measure the relative distributions of other folds than the largest. Hence, \dcp is a coarser measure that is used here for comparison purposes.

\subsection{Comparison of split quality measures}\label{ssec:ems}

Here we compare the behavior of the measures presented in Section~\ref{metrics} with a synthetic dataset. We generate different errors to class distribution with various splits of this data. Next, we show how different measures respond to these errors.

A good synthetic data should have the following charasteristics: 1) it should be large and sparse, 2) there should be high class imbalance. In order to achieve these conditions, the data was constructed as follows: The data consists of a binary target matrix $D \in \{0,1\}^{n\times q}$ with $n=100000, q=100$, that is, 100000 data points and 100 classes. In order to make the data class imbalanced, the positive class sizes of the data were set to vary from $2k$ to $n/2$, so that $|S_j| = 2k + \frac{j(n/2 -2k)}{q}, j \in 0\dots q-1$.
 
In order to ensure that the results are consistent, the synthetic data should be divided into cross validation folds that are of varying quality: 1) perfect (all class distributions are equal among folds), 2) with varying levels of error in the distributions between folds.

We divided the data $D$ into 10 cross validation folds in three ways: 1) all folds were equally sized, 2) there was a 20\% increase in one fold, 20\% decrease in one fold, and the rest were equal and 3) the class was missing from one fold while the other folds were equally sized. These are denoted as \text{Equal}, \text{Difference} and \text{One missing}, respectively. These three alternatives represent clearly different levels of error in data splitting, and measures should separate them from each other. Furthermore, all the classes of $D$ had the same fold distributions while the absolute class sizes were different.  Therefore, a good split quality measure should give a similar score for all classes, irrespective of their size. In practical settings it is especially important to correctly divide the smaller classes since the bigger classes are naturally better distributed.

\begin{figure}[H]
  \centering
    \subfloat[][\ld]{\includegraphics[width = 0.5\textwidth]{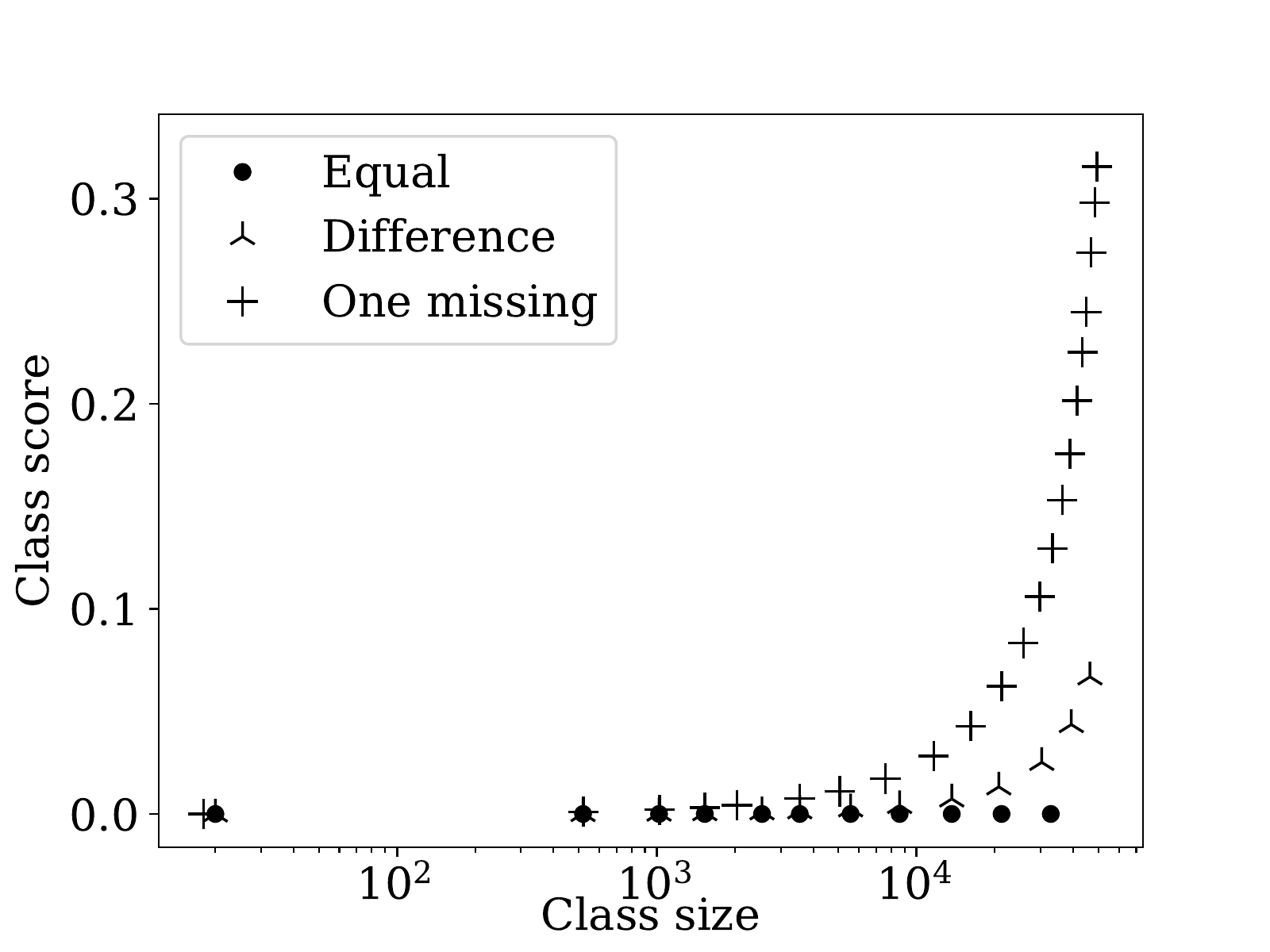}}
    \subfloat[][\wld]{\includegraphics[width = 0.5\textwidth]{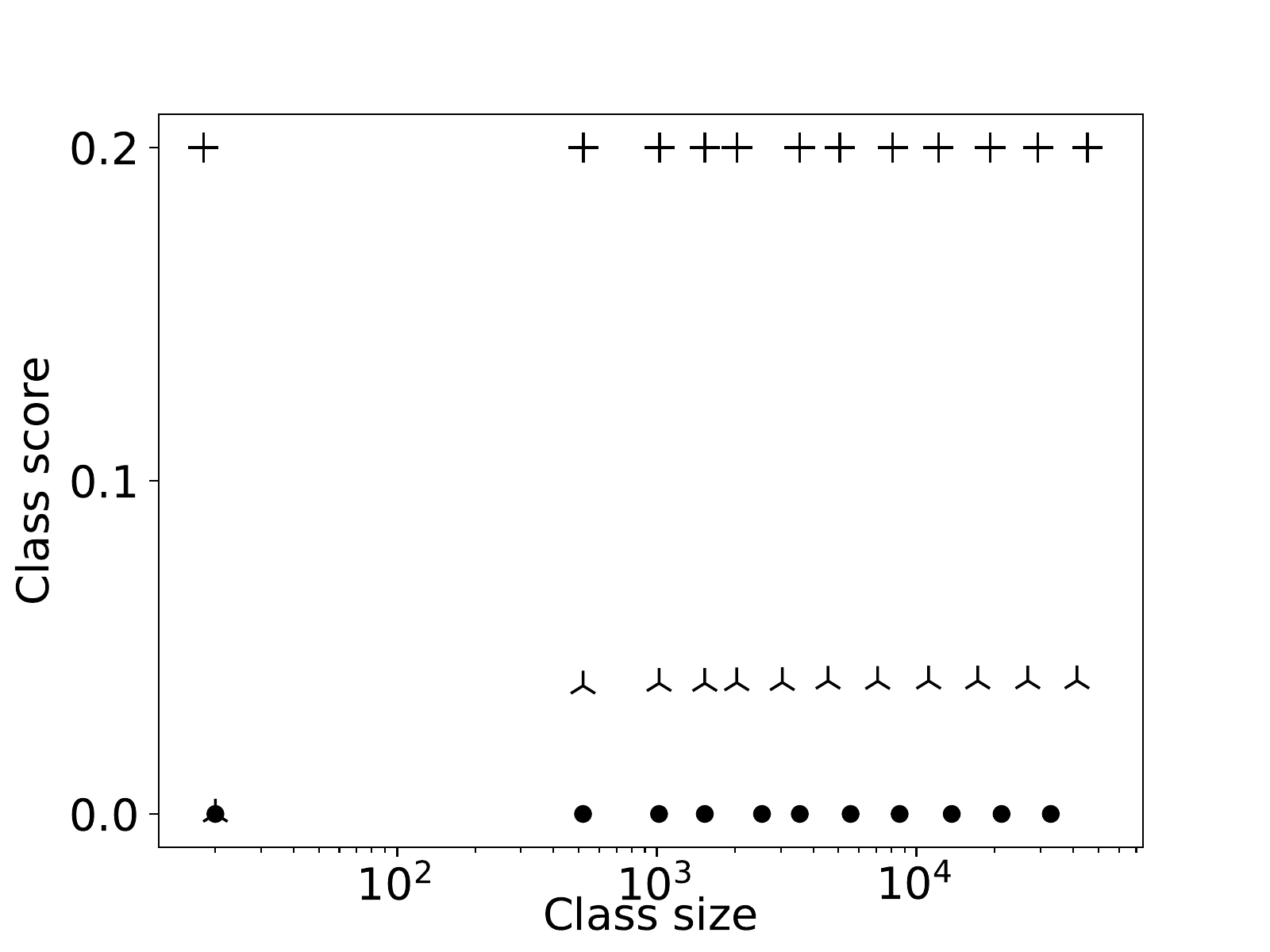}} \ \ \ \ 
    \subfloat[][\dcp]{\includegraphics[width = 0.5\textwidth]{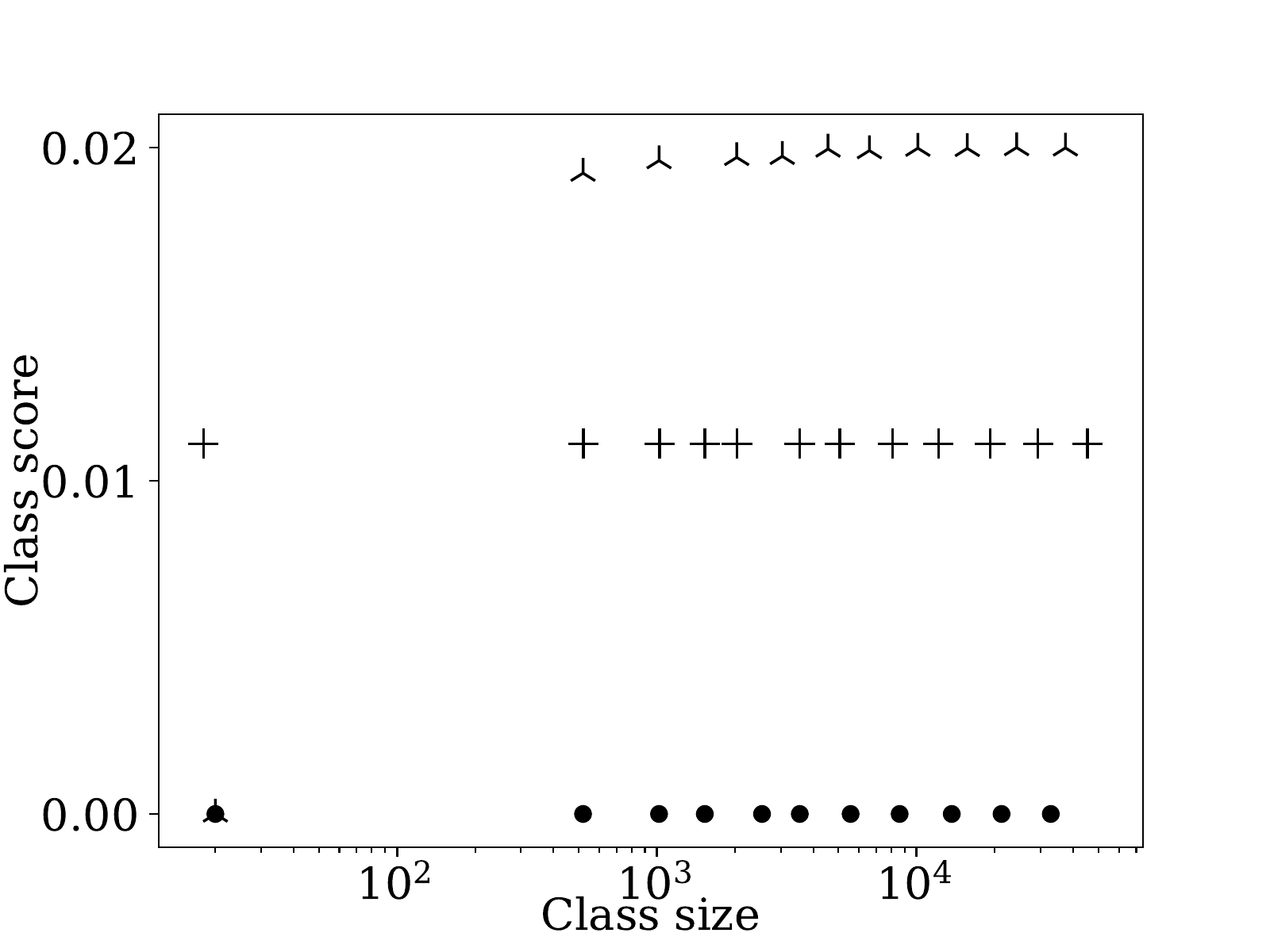}} 
  \caption{Behaviour of cross validation split quality measures on synthetic data with cross validation splits of varying quality. The cross validation folds have the same distribution for each class, so the scores should be equal. However, \ld score depends on the class size, while \wld and \dcp give the same weight for all classes. Note that as \dcp concentrates on the size of the largest fold, it behaves differently on data where the class is entirely missing from one fold giving it comparatively higher score than the other measures.}
    \label{synthetic}
\end{figure}

We evaluated these three splits splits on the synthetic data using the measures \ld, \wld and \dcp. The results for each measure are presented in the subplots of the Figure~\ref{synthetic}. In each plot, class size is plotted against the class specific score given by the measure (smaller is better) for the different folds. The results confirm that the widely used \ld depends on the class size. \ld gives smaller values to smaller classes and higher values to bigger classes even though their fold distributions are identical (see Figure~\ref{synthetic} (a)). The behaviour of \wld and \dcp on synthetic data, Figure~\ref{synthetic} (b) and (c), show that both \wld and \dcp are not affected by the class size. Therefore, it is recommendable to use them instead of \ld for measuring cross validation split quality.

\section{Algorithm}\label{sec:algorithm}

In this section, we present a new general algorithm, \own, for optimising cross validation splits with respect to any split quality measure that can produce class specific scores. \own can also be used to generate standard train-test splits by generating a cross validation split of size $n$ with fold size of the desired test set and then forming the training set from the folds $1\dots n-1$ and the test set from the fold $n$. Unlike some of the competing methods, \own does not need any data specific parameters that have to be adjusted for different datasets. The details of \own are presented formally in Algorithm~\ref{alg}.  

Intuitively, we start by randomly generating the initial $k$ subsets i.e. cross validation folds. Let $M$ be a measure that computes the quality of the folds with respect to a single class $c$. We calculate the score vector $M(c)$ for all classes $c\in C$ and calculate the global loss function $L = \sum_{c\in C}M(c)$.

Let $L_0$ be the initial value of $L$. Let $A$ be the index of the class with the highest $M$. We balance the class $A$ by moving data points from
folds with excess data points to folds without enough data points so that the class
distributions are balanced for the class $A$ (Function \texttt{balance} in Algorithm~\ref{alg}). After processing the class $A$ the
global loss $L$ is recalculated. Let $L_1$ be the recalculated loss. If $L_1 >
L_0$, undo the changes and move to the next worse class. Otherwise keep the
modification and continue to the next worse class. Continue this process for
all classes for many epochs until either the global loss does not improve any
more or a desired max iterations limit is reached. We only allow balancing
operations that lead to improvement in the global loss $L$. Balancing a class
can change the distribution of other classes. Thus it may be possible to
balance later a class that is skipped in the first epoch.

The time complexity of the Algorithm~\ref{alg} consists of processing $m$
classes, for each calculating the loss (complexity: $\bigohpar{nm}$, finding the
class with the highest loss ($\bigohpar{m}$) and redistributing the elements ($\bigohpar{n}$). Therefore, the total time complexity is $\bigohpar{nm^2}$.
In practice, \own could be also easily used on top of another possibly faster method to fine tune the results.

Note, that in the accompanying practical implementation the classes that have more positive than negative data points are balanced with respect to the negative distribution. This is not important for \go data but could be useful in some other applications. 

\begin{algorithm}[H]
  \SetAlgoLined
  \caption{Generating stratified cross validation splits} 
  \BlankLine
  \DontPrintSemicolon
  \SetKwProg{Fn}{Function}{}{}
  \SetKwInOut{params}{Input}\SetKwInOut{output}{Output}
  \SetKwFunction{balance}{balance}
  \SetKwFunction{stratifiedsplit}{stratified\_split}
  \SetKwFunction{randomsplit}{make\_random\_split}
  \params{measure $M$, \\ target data $D\in \{0,1\}^{n \times m}$, \\ number of cross validation folds $k$, \\ max number of epochs $\mathrm{max\_epochs}$}
  \output{List of cross validation splits}
\Fn{
    \stratifiedsplit{D, k}}{
        $\mathrm{folds}_0\leftarrow \randomsplit(D, k)$\;
    $\mathrm{epoch} \leftarrow 0$\;

    \While {true}{
        $\mathrm{max\_offset}\leftarrow 0$\;
        $L_0 \leftarrow M(D, \mathrm{folds}_0)$\;
       \For{$i \in [m]$}{  \label{algorithm}
           $A\leftarrow \argmax{L_0}[\mathrm{max\_offset}]$  \;
           $\mathrm{folds}_1 \leftarrow \balance(A)$ \;
           $L_1 \leftarrow M(D, \mathrm{folds}_1)$\;
         \If{$L_0 \leq L_1$}{
             $\mathrm{max\_offset}\leftarrow \mathrm{max\_offset} + 1$\;
            \Continue}

        $L_0 \leftarrow L_1$ \;
        $\mathrm{folds}_0 \leftarrow \mathrm{folds}_1$ \;
        }
        \If{$L_0 = l_1 \Or \mathrm{epoch} = \mathrm{max\_epochs}$}{
           \Break}
    }
    \Return{$\mathrm{folds}_0$}
  }
 \label{alg} 
\end{algorithm}

\section{Cross validation experiments}\label{sec:cvexp} 

In Section~\ref{metrics}, we showed that \ld is not a good measure for evaluating cross validation split quality. Most existing multilabel stratified cross validation methods found in the literature (see Section~\ref{related}) are optimised for producing splits with good \ld or similar class size dependent scores. In this Section, we will examine the performance of the existing methods, namely, \sois, \is, \sss and \pmbsrs as well as our own \own with respect to the proposed new measures \wld and \dcp. We will show that if one wants to get cross validation splits that are good with respect to \wld and \dcp, \own is the best option available since it can be used to directly optimise them.

In the following experiments we set $k=5$. All the results presented are averages over 10 runs with different random initialisations. The experiments were run using Python 3.6.8 on a machine with AMD opteron-6736 1.4GHz. Implementations of \own and the experiments presented in this article are available at \url{https://github.com/xtixtixt/optisplit}.

We used iterative stratification implementations from the popular \textit{skmultilearn} library \cite{skml} for first order and second
order iterative stratification experiments. The \textit{Stratified Sampling} \cite{merrillees21} implementation used was the one provided with the article. Note that the \sss implementation produces train test splits not cross validation splits. In order to compare it to the rest of the methods we have split the data by recursively splitting it to approximately 1/k sized test sets. Thus the method is run 5 times. 

The methods compared here can be divided into three categories. \is and \sois are iterative stratification based methods, \sss and \own are optimisation based methods and \pmbsrs is a random split based method.

We optimised \own with respect to both \wld and \dcp to compare the effect of the cost function to the outcome.

\subsection{Datasets}

We used a wide range of diverse datasets: \bibtex, \delicious and \mediamill from the MULAN dataset collection \cite{mulan} that have been used to evaluate earlier similar methods. Datasets \cc (cellular component), \mf (molecular function) and \bp (biological process) are our own \go subset datasets used in protein function prediction (see \cite{toronen2018pannzer2, zhou2019cafa} for more info). These and are considerably bigger and sparser than the MULAN datasets used here. The dataset \wiki \cite{Bhatia16}
is a large and very sparse extreme classification dataset.
Here, classes without any positive or negative data points are excluded. Detailed properties of the datasets are presented in Table~\ref{datasets}

\input{table1}

\input{table2}

\subsection{Results}
Scores of the split quality measures for all datasets and methods are presented in Table~\ref{results} with the following exceptions: \is and \sois results are not presented for the biggest datasets \bp and \wiki because their runtime was prohibitively high. \wiki results are not presented for \sss because the implementation used produced an error when run on that particular data. For comparison purposes we have also presented scores for random split (\random).

The results show that \own performs better than previous methods with respect to \wld and \dcp scores when optimising with either of those. The runtimes of \own are also competitive when compared to other top performing methods. Generally, iterative stratification based methods perform quite well but are unusably slow on bigger datasets. Random split based methods are fast but produce poor quality folds compared to more advanced methods. Optimisation based methods (\own and \sss) usually give best results and their runtimes are in the middle of iterative stratification based and random split based methods.

We can see that \dcp and \wld are very correlated, the ordering of the methods is similar with respect to both measures and optimising \own with respect to \dcp produces nearly as good \wld results as optimising \wld directly.
However, since \wld measures the folds more thoroughly i.e. it does not just concentrate on the biggest fold it seems to be a better practical choice than \dcp in most cases.

Note that \own does not attempt to produce exactly equally sized splits. This results in quite high \ed (Example Distribution) scores compared to some other methods. This should not be a problem in practical machine learning settings. 

For completeness, we have included \ld evaluations in Table~\ref{results}. As is to be expected from the results presented in Section~\ref{metrics}, we can see that the method ordering is often considerably different with respect to \ld scores. In smaller and less imbalanced datasets \ld gives results more in line with \wld and \dcp. For bigger and more imbalanced datasets, when \ld weakness gets more pronounced, the results differ more significantly. There, \ld favours iterative stratification based methods and gives random split based \pmbsrs noticeably better score than to \random, in contrast to \wld or \dcp evaluations.

\section{Discussion and future work}\label{discussion}
In this article, we have shown that the most widely used multilabel cross validation split quality measure, \ld, does not measure split quality correctly when used on unequally sized classes. In response, we have presented new measures with better properties and have presented a new general method, \own, for generating and optimising multilabel stratified cross validation splits. We have compared \own to existing methods and found that it produces better quality cross validation folds with respect to the new measures than the previous methods and scales well for \go sized datasets. We note for future work that \own could be made faster by calculating the loss only for the classes that have been modified in the previous balancing operation. For sparse data that should allow it to be used even on considerably larger datasets. Also, \own uses now a greedy hill-climbing approach for optimising the target function. However, a Monte Carlo / simulated annealing based version could achieve even better performance. 

\section*{Acknowledgements}
This work was funded by NNF20OC0065157 of the Novo Nordisk Foundation. Computations were partly done using resources belonging to Biocenter Finland's Bioinformatics platform.

\bibliographystyle{plain}
\bibliography{bibliography.bib}
\end{document}

%% file: table1.tex
\begin{table}
    \begin{tabular}{|c|c|c|c|c|c|c|c|c|}
    \hline
    Data & Data size & Labels & Density & Min & 25\% & 50\% & 75\% & Max \\ \hline
    \bibtex & 7395 & 159 & 0.0151 & 51 & 61 & 82 & 130 & 1042 \\ \hline
    \delicious & 16015 & 983 & 0.0193 & 21 & 58 & 105 & 258 & 6495 \\ \hline
    \mediamill & 43907 & 101 & 0.0433 & 31 & 93 & 312 & 1263 & 33869 \\ \hline
    \cc & 577424 & 1688 & 0.0077 & 5 & 66 & 225 & 891 & 577410 \\\hline
    \mf & 637552 & 3452 & 0.0028 & 11 & 61 & 150 & 498 & 637533 \\ \hline
    \bp & 666349 & 11288 & 0.0028 & 4 & 41 & 123 & 493 & 666338\\ \hline
    \wiki & 20762 & 30938 & 0.0006 & 2 & 2 & 3 & 6 & 16756 \\  \hline
    \end{tabular}
    \caption{Properties of the datasets used in evaluations. \cc, \mf and \bp are \go subsets. Min and Max represent the minimum and maximum class sizes of the datasets. Columns 25\%, 50\% and 75\% are the corresponding percentiles of the class sizes.}
     \label{datasets} 
\end{table}

%% file: table2.tex
\begin{table}
\vspace{-42pt}
\begin{tabular}{|c|c|r|r|r|r|r|}
    \hline
    Dataset & Method & \ed & \ld & \dcp & \wld & Runtime (s)  \\ \hline

    \multirow{4}{*}{\bibtex} & $\own_\wld$ & 27 & 0.0004 & 0.0073 & \textbf{0.0234} & 5 \\
                             & $\own_\dcp$ & 38 & 0.0005 & \textbf{0.0068} & 0.0315 & 5\\
                             & \sois & 16 & 0.0005 & 0.0143 & 0.0425 &  5 \\
                             & \is & 17 & 0.0007 & 0.0206 & 0.0604 &  1 \\
                             & \sss & 57 & 0.0007 & 0.0173 & 0.0465 & 4 \\
                             & \random & 0 & 0.0022 & 0.0564 & 0.1693 & 1 \\
                             & \pmbsrs & 2 & 0.0022 & 0.0558 & 0.1660 & 1 \\ \hline

    \multirow{4}{*}{\mediamill} & $\own_\wld$ & 53 & 0.0005 & 0.0053  & 0.0187 &  10 \\
                                & $\own_\dcp$ & 7 & 0.0005 & \textbf{0.0047} & \textbf{0.0176} &  11 \\
                                & \sois & 1 & 0.0003 & 0.0205 & 0.0610 &  77 \\
                                & \is & 1 & 0.0008 & 0.0280 & 0.0854 &  50 \\
                                & \sss & 36 & 0.0009 & 0.0068 & 0.0231 &  31 \\
                                & \random & 0 & 0.0019 & 0.0379 & 0.1142 & 1 \\
                                & \pmbsrs & 2 & 0.0019 & 0.0386 & 0.1150 & 1 \\ \hline

    \multirow{4}{*}{\delicious} & $\own_\wld$ & 35 & 0.0010 & 0.0223 & 0.0666 &  75 \\
                                & $\own_\dcp$ & 28 & 0.0010 & \textbf{0.0215} & 0.0772 &  76 \\
                                & \sois & 16 & 0.0012 & 0.0458 & 0.1357 & 381 \\
                                & \is & 13 & 0.0013 & 0.0489 & 0.1461 & 11 \\
                                & \sss & 75 & 0.0007 & 0.0221 & \textbf{0.0625} &  40 \\
                                & \random & 0 & 0.0015 & 0.0507 & 0.1515 & 1 \\
                                & \pmbsrs & 2 & 0.0015 & 0.0512 & 0.1525 & 1 \\ \hline

    \multirow{4}{*}{\cc} & $\own_\wld$ & 193 & 8.1053 & 0.0065 & \textbf{0.0230} & 2762 \\
                        & $\own_\dcp$ & 78 & 8.1195 & \textbf{0.0062} & 0.0248 &  2872 \\
                         & \sois & 5 & 5.7623 & 0.0305 & 0.0894&  204823 \\
                        & \is & 1 & 5.6075 & 0.0448 & 0.1320&  97120 \\
                         & \sss & 182 & 8.8831 & 0.0133 & 0.0416&  1118 \\
                        & \random & 0 & 10.2802 & 0.0455 & 0.1342 &  1 \\
                        & \pmbsrs & 1 & 6.3606 & 0.0448 &  0.1332 &  17 \\ \hline

    \multirow{4}{*}{\mf} & $\own_\wld$ & 607 & 3.2147 & 0.0064 & \textbf{0.0229} & 4980 \\
                        & $\own_\dcp$ & 89 & 3.1782 & \textbf{0.0063} & 0.0252 &  5209 \\
                         & \sois & 84 & 3.0536 & 0.0490 & 0.1450&  53646 \\
                         & \is & 1 & 2.7503 & 0.0490 & 0.1451&  20442 \\
                         & \sss & 656 & 4.7935 & 0.0129 & 0.0400 &  1018 \\
                         & \random & 0 & 6.2255 & 0.0493 & 0.1465 &  1 \\
                         & \pmbsrs & 1 & 4.8213 & 0.0498 & 0.1480 &  17 \\ \hline

    \multirow{4}{*}{\bp} & $\own_\wld$ & 612 & 2.1053 & 0.0161 & \textbf{0.0516} & 59436 \\
                         & $\own_\dcp$ & 173 & 2.0951 & \textbf{0.0156} & 0.0576 & 52412 \\
                         & \sss & 279 & 2.9140 & 0.0282 & 0.0857 &  2734 \\
                         & \random & 0 & 2.896 & 0.0568 & 0.1664 &  1 \\
                         & \pmbsrs & 0 & 2.4250 & 0.0567 & 0.1662 &  21 \\ \hline

    \multirow{4}{*}{\wiki} & $\own_\wld$ & 1678 & 0.0002 & 0.2579 & \textbf{0.7563} & 3065 \\
                         & $\own_\dcp$ & 367 & 0.0002 & \textbf{0.2068} & 0.8033 & 3053 \\
                           & \sss & error & error & error & error & error \\
                           & \random & 0 & 0.0002 & 0.3008 & 0.9316 &  1 \\
                           & \pmbsrs & 2 & 0.0002 & 0.3013 & 0.9323 & 1 \\ \hline

\end{tabular}
\caption{Performances of the algorithms evaluated on diverse datasets. The scores are means over 10 runs. Bold font highlights the best DCP and rLD results for each dataset. Error marks case where the method failed to run. Notice that we ran slower methods (SOIS and IS) only with smaller datasets.}
 \label{results} 
\end{table} 